\documentclass[letterpaper, 10 pt, conference]{ieeeconf}  

\IEEEoverridecommandlockouts                              

\title{\LARGE \bf
Imitation Learning-Based Online Time-Optimal Control with Multiple-Waypoint Constraints for Quadrotors}
\usepackage{graphicx}
\usepackage{stfloats}
\usepackage{placeins}
\usepackage{subfigure}
\usepackage{amsmath}
\usepackage{mathtools}
\usepackage{float}
\usepackage{amsfonts,amssymb}
\usepackage{multirow}
\usepackage{makecell}
\usepackage{tabulary}
\usepackage{hyperref}
\hypersetup{hidelinks,
	colorlinks=true,
	allcolors=blue,
	pdfstartview=Fit,
	breaklinks=true}

\author{Jin Zhou$^{1}$, Jiahao Mei$^{2}$, Fangguo Zhao$^{1}$, Jiming Chen$^{1}$, \IEEEmembership{Fellow,~IEEE}, and Shuo Li$^{1}$
\thanks{$^{1}$Authors are with the College of Control Science and Engineering, Zhejiang University, Hangzhou 310027, China
        {\tt\small shuo.li@zju.edu.cn}
        }%
\thanks{$^{2}$Jiahao Mei is with the Department of Automation, Zhejiang University of Technology, Hangzhou 310023, China.
        }%
}

\begin{document}

\maketitle
\thispagestyle{empty}
\pagestyle{empty}


\begin{abstract}
Over the past decade, there has been a remarkable surge in utilizing quadrotors for various purposes due to their simple structure and aggressive maneuverability, such as search and rescue, delivery and autonomous drone racing, etc. One of the key challenges preventing quadrotors from being widely used in these scenarios is online waypoint-constrained time-optimal trajectory generation and control technique. This letter proposes an imitation learning-based online solution to efficiently navigate the quadrotor through multiple waypoints with time-optimal performance. The neural networks (WN\&CNets) are trained to learn the control law from the dataset generated by the time-consuming CPC algorithm and then deployed to generate the optimal control commands online to guide the quadrotors. To address the challenge of limited training data and the hover maneuver at the final waypoint, we propose a transition phase strategy that utilizes MINCO trajectories to help the quadrotor 'jump over' the stop-and-go maneuver when switching waypoints. Our method is demonstrated in both simulation and real-world experiments, achieving a maximum speed of $5.6m/s$ while navigating through $7$ waypoints in a confined space of $5.5m \times 5.5m \times 2.0m$ [\href{https://youtube.com/playlist?list=PLrUtAo-zaR9xmOzuae2mXLVsNZxzyAgWa\&si=1WNQfKzKF6gG36dK}{\textbf{\emph{video}}}\footnote[3]{https://youtube.com/playlist?list=PLrUtAo-zaR9xmOzuae2mXLVsNZxzyAgWa\&si=1WNQfKzKF6gG36dK}]. The results show that with a slight loss in optimality, the WN\&CNets significantly reduce the processing time and enable online optimal control for multiple-waypoint constrained flight tasks.
\end{abstract}

\hypersetup{hidelinks,
	colorlinks=true,
	allcolors=black,
	pdfstartview=Fit,
	breaklinks=true}

\section{INTRODUCTION}
Quadrotors are one of the most agile unmanned aerial vehicles, which have immense potential for time-critical tasks such as delivery, inspection, search and rescue operations, etc \cite{hanover2024autonomous}. Over the past decade, many researchers have been dedicated to addressing the challenges in navigation, trajectory generation and control, aiming to push the boundaries of quadrotors to execute aggressive flight in real-world applications \cite{zhou2019robust,ren2022bubble}.
\begin{figure}[hbt]
    \centering
    \includegraphics[width=0.45\textwidth,trim=350 180 350 150,clip]{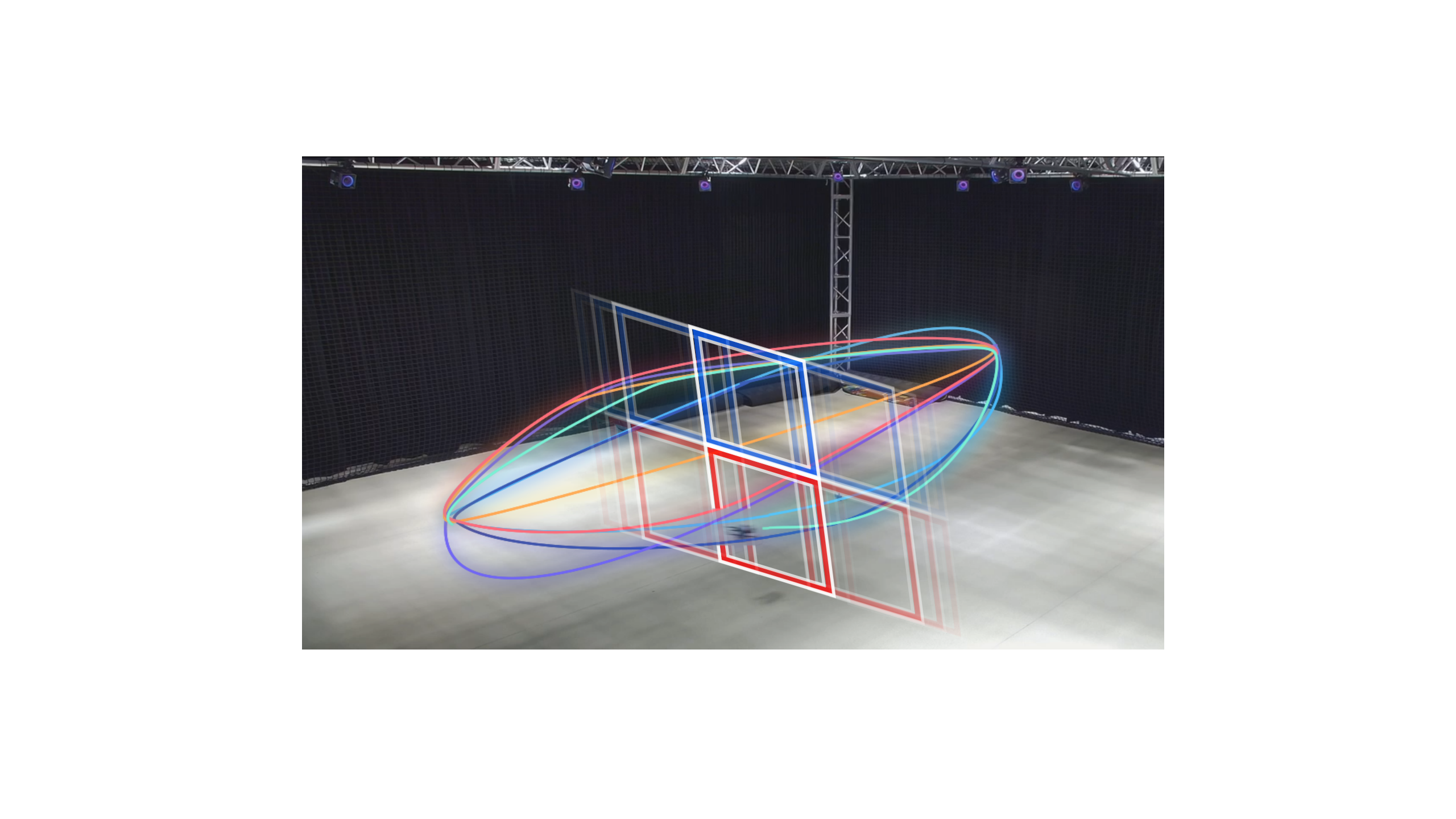}
    \caption{The quadrotor continuously passes through moving target points at high agility with the proposed WN\&CNets and the MINCO transition.}
    \vspace{-0.35cm}
    \label{fig: exp4}
\end{figure}
In this letter, we discuss the scenario where the quadrotor has to autonomously fly through multiple waypoints (including dynamic waypoints whose trajectories are unknown in advance) aggressively with time-optimal trajectories in a confined space, e.g. drone racing challenges \cite{foehn2022alphapilot} and indoor rescue \cite{cui2016search}. We present an imitation learning-based approach and develop a neural network called WN\&CNets (waypoint-constrained navigation and control network) to generate control commands online given the waypoints and the states of the quadrotor. Our approach can make quadrotors replan in real-time with high frequency to handle multiple waypoints including dynamic waypoints and also the deviation caused by disturbance or model inaccuracy.

Common approaches for aggressive trajectory generation can generally be classified into three categories: polynomial-based methods, sampling-based methods and optimization-based methods. Polynomial-based methods \cite{mellinger2011minimum, bry2015aggressive} leverage the quadrotors' differential flatness property \cite{faessler2017differential} to efficiently calculate trajectories subject to waypoint constraints. However, due to the inherent smoothness of polynomials, quadrotors are unable to reach their input limits consistently with polynomial trajectories at finite order or with finite segments of polynomials \cite{hanover2024autonomous}. Sampling-based methods \cite{webb2013kinodynamic,liu2018search} search in the state spaces until available trajectories are found or specific trajectories are considered and the best one is selected. However, the high dimensionality of quadrotors' state space poses a challenge when conducting these searches within the limited time. Simplifying the quadrotor's dynamic model is a common approach to increase the solving efficiency but it will lose part of the quadrotors' maneuverability and the time optimality. Optimization-based methods emerge as a viable approach to avoid sacrificing optimality. State-of-the-art optimization algorithms such as the CPC \cite{foehn2021time}, are capable of calculating optimal trajectories for a quadrotor navigating through multiple waypoints and can be extended to multi-drone scenarios \cite{shen2023aggressive}. However, these optimization algorithms typically rely on numerical methods, which can be time-consuming. Thus, they are impractical for situations where online calculation and replanning are needed, such as aggressive flight in dynamic environments, etc. 

To achieve online time-optimal flight with multiple-waypoint constraints, various innovative planning and control methods are developed and combined for autonomous drone systems \cite{wang2022geometrically,fork2023euclidean,qin2023time}. Model Predictive Contouring Control (MPCC) \cite{romero2022model,romero2022time} is developed to track pre-defined trajectories while providing time-optimal online replanning. Deep reinforcement learning (RL) \cite{song2023reaching,kaufmann2023champion} is utilized to train control policy for navigating racing tracks while coping with model uncertainty. Imitation learning also presents an alternative online approach by using neural networks to learn optimal trajectories \cite{tang2018learning,izzo2020real,li2020aggressive}. However, these works do not consider multiple waypoints or are limited to two-dimension spaces. Recently, the Guidance \& Control Networks \cite{origer2023guidance} have been trained for tasks involving two upcoming waypoints, but the waypoints' position is very limited and there are relatively large errors when flying through the waypoint.



\begin{figure*}[hbt]
    \centering
    \vspace{-0.35cm}
    \includegraphics[width=0.96\textwidth, trim = 230 160 140 170, clip]{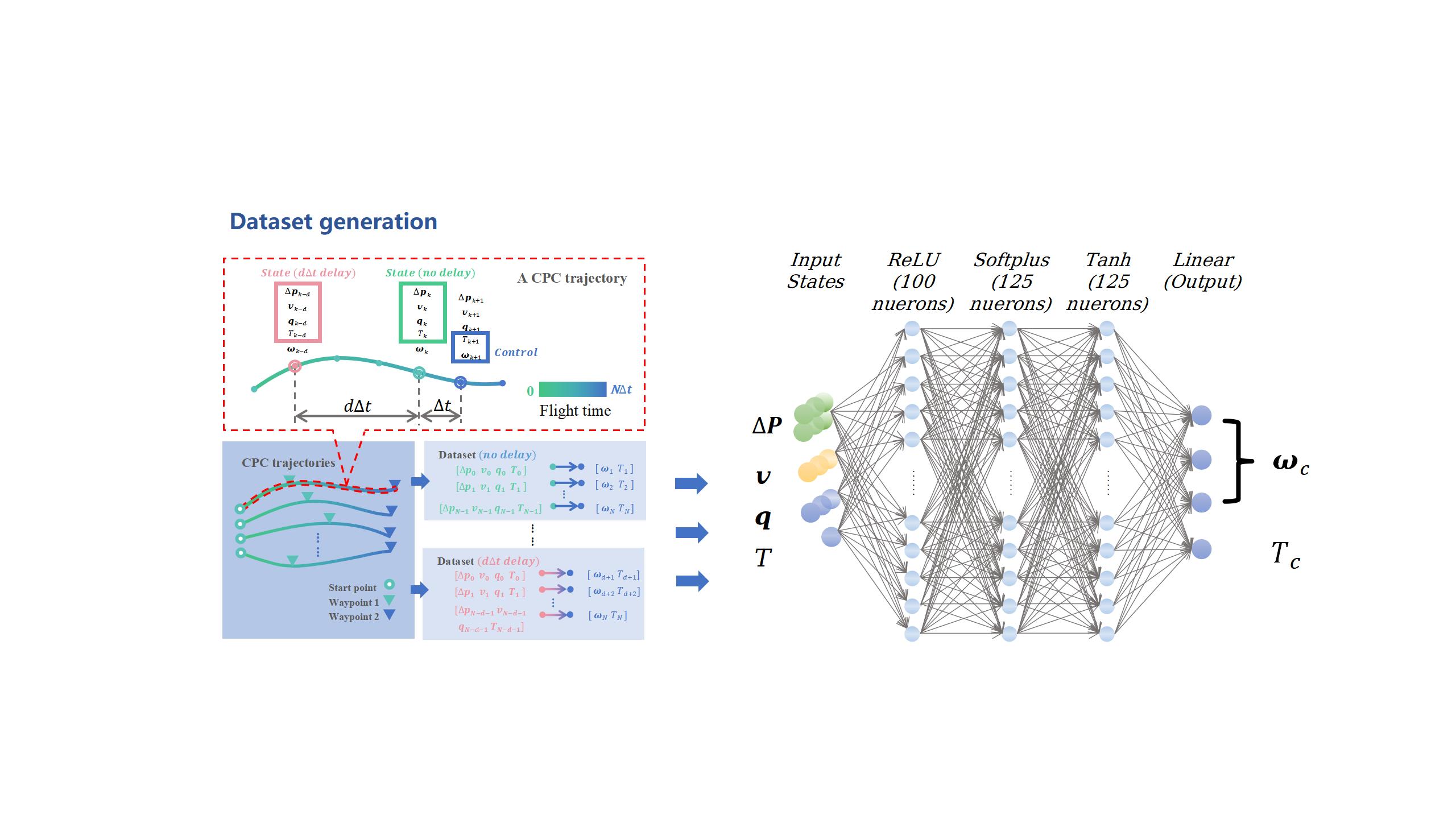}
    \caption{Datasets that contain a great amount of state-control pairs are generated using the CPC method. These state-control pairs are then taken as training data for WN\&CNets. The WN\&CNets will then be deployed to guide the quadrotor to fly through multiple waypoints.}
    \label{fig: core}
\vspace{-0.35cm}
\end{figure*}

In this letter, we propose a novel neural network, the WN\&CNets, to learn the control law in completely time-optimal trajectories with $2$ waypoints, which are generated by the CPC method \cite{foehn2021time} (Fig. \ref{fig: core}). Based on the original WN\&CNets, we also apply MINCO trajectories \cite{wang2022geometrically} as a lightweight tool to develop a transition phase strategy to help the quadrotor 'jump over' the stop-and-go maneuver introduced by the WN\&CNets, due to the limited computing resources. With the aid of the MINCO transition, the WN\&CNets can empower the quadrotor to consecutively fly through multiple waypoints in a confined space, performing remarkably aggressive and agile flight maneuvers without hovering. 

Hence, in this letter, we 

 \begin{enumerate}
        \item adapt the CPC method to generate a dataset containing state-control pairs of the time-optimal trajectories and develop WN\&CNets (Waypoint-constrained Navigation and Control Networks) to learn the optimal control law in the dataset. Deploy the WN\&CNets to generate aggressive control commands online.  
        \item propose a MINCO-based transition phase strategy to help the quadrotor 'jump over' the stop-and-go maneuver when switching waypoints while ensuring that the quadrotor can keep near time-optimal performance with very limited training data.  
	    \item validate and analyze our approach both in simulation and real-world flight tests where the quadrotor navigates through $7$ waypoints with a maximum speed of $5.6m/s$ in a space of $5.5m \times 5.5m \times 2.0m$ and fly with complicated trajectories (up to 22 waypoints) to demonstrate the feasibility of the proposed method. 
\end{enumerate}

\section{METHODOLOGY}
\subsection{Quadrotor dynamics}
To generate the time-optimal dataset, we first describe the quadrotor's model we use in this letter. The states of a quadrotor are
\vspace{-0.35cm}
\begin{align}
{\mathbf{x}} = [\mathbf{p}, \mathbf{v}, \mathbf{q}, \boldsymbol{\omega}]
\end{align}
In the equations above ${\mathbf{p}}$, ${\mathbf{v}}$, ${\mathbf{q}}$ and ${\boldsymbol{\omega}}$ represent position, velocity, attitude and angular velocity, respectively.

In this letter, we utilize the same quadrotors' dynamics model as the one previously presented in the work of \cite{foehn2021time}, which includes a linear drag term \cite{faessler2017differential} represented by an approximated diagonal matrix $\mathbf{D}=diag(d_x,d_y,d_z)$. 
\begin{align}
{\dot{\mathbf{x}}} = \mathbf{f}_{dyn}({\mathbf{x}},{\mathbf{u}}) =
\begin{cases}
{\mathbf{v}} \\
\mathbf{g}+\frac{1}{{m}}\mathbf{R}({\mathbf{q}}){{\mathbf{u}}} - \mathbf{R}({\mathbf{q}})\mathbf{D}\mathbf{R}^T({\mathbf{q}})\mathbf{v} \\
\frac{1}{2}\Lambda({\mathbf{q}})
\begin{bmatrix}
0 \\ {\boldsymbol{\omega}} 
\end{bmatrix} \\
 {\mathbf{J}}^{-1}( {\boldsymbol{\tau}} - {\boldsymbol{\omega}}\times {\mathbf{J}}{\boldsymbol{\omega}})
\end{cases}
\label{equ:quadrotor dynamics}
\end{align}

Readers are referred to \cite{foehn2021time} for the details of the meaning of the symbols.


\subsection{The optimization problem}
\label{sec:optimization problem}
Traditional approaches for solving planning problems with multiple waypoints typically rely on having prior knowledge of the time allocation for each individual waypoint. However, reconciling the total trajectory time with the time allocation for waypoints poses a conflicting optimization challenge. To address this, the method utilizing the complementary progress constraints (CPC) \cite{foehn2021time} has emerged. This method enables the representation of constraints through waypoints, overcoming the conflict between trajectory time and waypoint time allocation.

The CPC problem for a specific planning task is formulated on a set of discrete time segments, with the number of the time segments $N$ determined based on the estimation of the flight task distance. The system states include dynamic states and progress states, which can be presented as
\begin{align*}
\centering
    {\mathbf x}_k &= \begin{bmatrix}
        {{\mathbf{p}}}_k & {{\mathbf{v}}}_k & {{\mathbf{q}}}_k & 
       {{\boldsymbol{\omega}}}_k & {{\mathbf{u}}}_k & {\boldsymbol{\lambda}_k} & {\boldsymbol{\mu}_k} & {\boldsymbol{\nu}_k} 
    \end{bmatrix}
\end{align*}

As minimizing flight time is the only goal of the optimization problem, the problem can be described as 
\begin{align}
{\mathbf{X}^*} = arg\min\limits_{X} t_N
\end{align}
where $\mathbf{X}$ is the set of optimization variables in this problem, including total time $t_N$ and the system state at each node
\begin{align*}
{\mathbf{X}} = [t_N, \mathbf{x}_0, \mathbf{x}_1, ..., \mathbf{x}_{N}]
\end{align*}

The constraints of the optimization include dynamic constraints, boundary constraints, input constraints, subsequence constraints and complementary progress constraints, which are based on the original CPC method \cite{foehn2021time}. For the complementary progress constraints,


\vspace{-0.35cm}
    \begin{align}
        \begin{cases}
             {\mu_{k}^j}(\left \Vert {\mathbf{P}_k} - {\mathbf{P}}_{wj} \right \Vert_2^2-{\nu}_k^j) := 0 \\
             -\nu^j_k \leq 0 \\
             \nu^j_k - d^2_{tol} \leq 0 
        \end{cases}
        \label{cpc:origin}
    \end{align}
    where $d_{tol}$ is the defined upper bound of the tolerance range, and ${\mathbf{P}}_{wj}$ is the position of the $j^{th}$ waypoint. If we only consider the time optimality without heading constraints, we use the same constraints (\ref{cpc:origin}) as in \cite{foehn2021time}. However, for some flight tasks where the heading at the waypoint needs to be specified, we will revise constraint (\ref{cpc:origin}) to
    \begin{align}
        \begin{cases}
             {\mu_{k}^j}(\left \Vert {\mathbf{P}_k} - {\mathbf{P}}_{wj} \right \Vert_2^2 + (\psi_k-\psi_{wj})^2 -{\nu}_k^j) := 0 \\
             -\nu^j_k \leq 0 \\
             \nu^j_k - d^2_{tol} \leq 0 
        \end{cases}
        \label{cpc:yaw}
    \end{align}
    where $\psi_k$ is the heading at node $k$ and $\psi_{wj}$ is the desired heading at the $j^{th}$ waypoint. In this letter, we focus on generating trajectories which are as fast as possible, and thus do not impose any constraints on the heading. However, in the next section, some experimental attempts and analysis are conducted on using heading as one of the target states for waypoints.
    

We use the CasADi toolkit and IPOPT solver to solve the optimization problem with predefined waypoints and boundary conditions. Readers are referred to \cite{foehn2021time} for the details of the CPC method. Subsequently, we will obtain a time-optimal trajectory consisting of a series of discrete nodes.
                
\subsection{Dataset generation}
The computational complexity of solving the optimization problem above makes it impossible to run the solver online during the flight. To tackle this problem, we use the WN\&CNets to map the quadrotor's waypoints, current states and the next step states to form an online controller. However, to generate the dataset, traversing all the states and the waypoints becomes impossible with the increasing quadrotor's state space and the number of waypoints. Thus, with limited computational resources, to generate the dataset for the WN\&CNets, in this letter, we only generate the time-optimal trajectories from hover states to hover states while passing through $2$ waypoint (the latter one is a hover state waypoint). Each time-optimal trajectory has $N+1$ nodes and node $k(k \in [0, N])$ contains the quadrotor's state at $t_k$, including relative position to the waypoints $\Delta \mathbf{p}_k$, velocity $\mathbf{v}_k$, attitude $\mathbf{q}_k$, rate $\boldsymbol{\omega}_k$, and thrust $T_k$. Notably, the original CPC method \cite{foehn2021time} uses quaternions to represent attitudes and provides a general acceleration as the thrust command. However, the relationship between quaternions and the control commands exhibits a high degree of nonlinearity, which poses challenges during the training process. Instead of quaternions, we use roll, pitch and yaw to represent the orientation in this work.

To be more specific, we can take $\Delta \mathbf{p}_k$, velocity $\mathbf{v}_k$, attitude $\mathbf{q}_k$ and thrust $T_k$ as the inputs of the network and $\boldsymbol{\omega}_{k+1}$ and thrust $T_{k+1}$ as the output of the neural network to form the state-control pairs, that is 
\begin{align}
\begin{bmatrix}
    \Delta \mathbf{p}_k & \mathbf{v}_k & \mathbf{q}_k & T_k
\end{bmatrix} \rightarrow 
\begin{bmatrix}
    \boldsymbol{\omega}_{k+1} & T_{k+1}
\end{bmatrix}
\end{align}


However, in real world, there are delays from state observations, onboard processing time, communications between the quadrotor and the ground control station, etc. If the step time is $\Delta t$ and the delay duration is $d\Delta t (d=0,1,2,3,...)$, we need to adjust our state-control pairs to account for those delays. Thus, instead of using the states of node $k$ as input, we use the state of node $(k-d)$ as the inputs of the WN\&CNets to predict the optimal states at $t_{k+1}$, that is
\begin{align}
\begin{bmatrix}
        \Delta \mathbf{p}_{k-d} & \mathbf{v}_{k-d} & \mathbf{q}_{k-d} & T_{k-d}
    \end{bmatrix} \rightarrow 
    \begin{bmatrix}
        \boldsymbol{\omega}_{k+1} & T_{k+1}
    \end{bmatrix}
\end{align}
This method yields $N-d$ state-control pairs as training data from one time-optimal trajectory. In our case, we generate $37044$ time-optimal trajectories and each trajectory has about $80$ nodes (the number of nodes depends on the length of the trajectory). So we have around $3,000,000$ state-control pairs to form the dataset. 
\subsection{WN\&CNet architecture and training}
The inputs for the WN\&CNets consist of the quadrotor's relative positions to the waypoints ($\Delta \mathbf{p} \in \mathcal R^{3\times 2}$), velocity ($\mathbf{v} \in \mathcal R^3$), attitude ($\mathbf{q} \in \mathcal R^3$), and thrust ($T$) of node $k-d$. The corresponding outputs of the WN\&CNets are the angular rates ($\boldsymbol{\omega}_c \in \mathcal R^3$) and thrust commands ($T_c$) of node $k+1$ which will be sent to the quadrotors.


The general network architecture employed in this work is depicted in Fig. \ref{fig: core}. For the training process, we utilize the mean squared error (MSE) as the loss function. We employ the Adam optimizer to update the weights of the networks. We build the network and conduct training using the PyTorch framework. The loss decreases quickly during the training progress and reaches about $0.04$ after 40 epochs. 

\subsection{MINCO-based transition phase strategy}
As stated above, to decrease the computational burden, in our dataset, we only have $2$ waypoints and the second one is a hover-state waypoint. So the quadrotor needs to hover at the second waypoint. Thus, for a multiple-waypoint flight task, a waypoint-switching strategy is needed to switch the waypoint during the flight. An intuitive waypoint-switching strategy is that when the quadrotor arrives at the second waypoint, it switches the next two waypoints as its target resulting in a stop-and-go maneuver, which violates the time-optimal purpose. 

To solve this issue, we need a transition phase to a) help the quadrotor jump over the stop-and-go maneuver, b) keep near time-optimal performance, and c) run online. In this work, we choose MINCO trajectories \cite{wang2022geometrically} to help the quadrotor 'jump over' the stop-and-go maneuver. MINCO is a novel minimum control effort optimization framework with efficient computation. By parameterizing the quadrotor’s flat output trajectory, it can eliminate constraints on system dynamics, initial state, and final state  while ensuring precise waypoint traversal. We apply it as a lightweight tool to compute polynomials while minimizing traversal time. We directly give the waypoints and don't need to construct safe flight corridors (SFC) in the framework. 


\begin{figure}[!h]
    \centering
     \subfigure[For same direction flights, the velocity is used to determine the entrance and exit of the transition phase. MINCO trajectories are used to jump over the stop-and-go maneuver.]{\includegraphics[width=0.47\textwidth, trim = 180 150 230 150, clip]{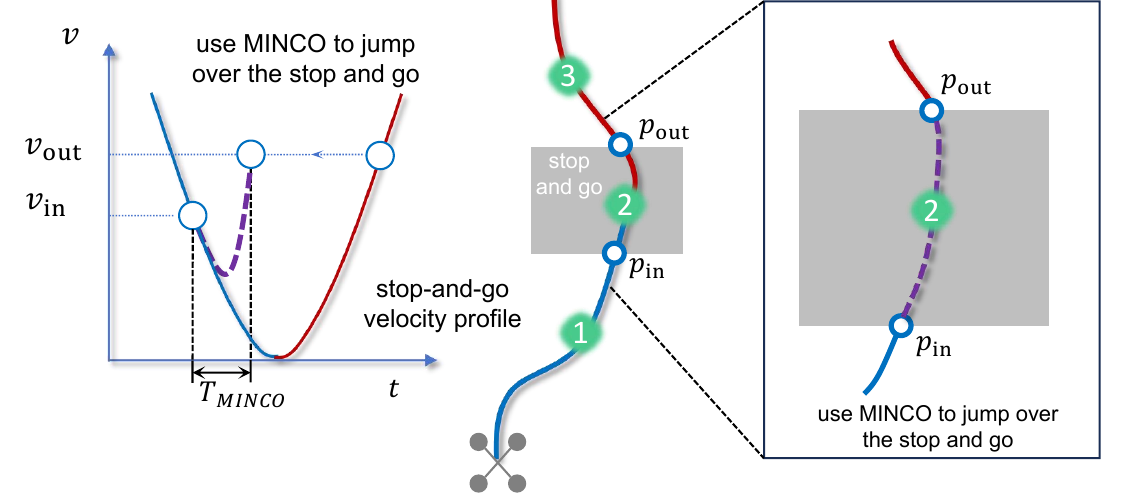}} 
     \subfigure[For opposite direction flights, use acceleration to determine the entrance and exit of the transition phase.]{\includegraphics[width=0.47\textwidth, trim = 180 150 220 150, clip]{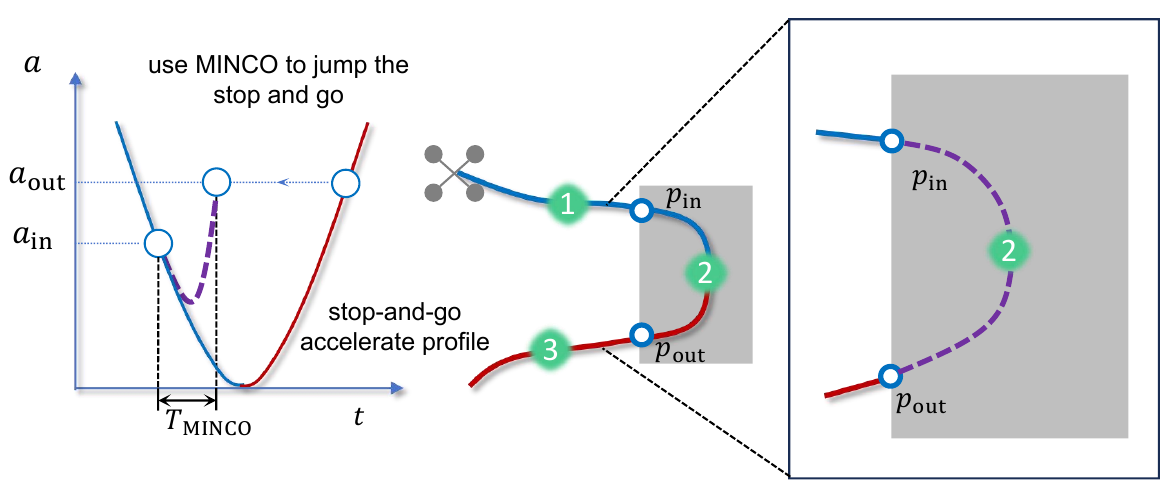}}
    \caption{Illustrative sketch of the CPC and its deficiency in multi-drone racing scenarios.}
    \vspace{-0.25cm}
    \label{fig:minco sketch} 
\end{figure}
An important challenge lies in determining when to enter and exit the transition phase. We discuss our strategy in $2$ scenarios. Let $\mathbf{wp}_i$ ($i=1,2,3$) represent the position vector of the $i^{th}$ waypoint. The first scenario is shown in Fig. \ref{fig:minco sketch} (a) where $(\mathbf{wp}_2-\mathbf{wp}_1)(\mathbf{wp}_3-\mathbf{wp}_2)>0$, which we call 'same direction flight'. After waypoint 1, with the WN\&CNets, the quadrotor will decelerate to hover at waypoint 2 and then accelerate to waypoint 3. Its velocity curve resembles a 'v' shape which leads to a stop-and-go maneuver. To avoid this maneuver, we set a parameter $v_{in}$ as a threshold to determine the entrance of the transition phase and we denote the corresponding point by $p_{in}$. Similarly, we set another threshold $v_{out}$ to determine the exit of the transition phase and the corresponding point is denoted by $p_{out}$. At the point where the predicted velocity of the quadrotor is larger than the threshold $v_p > v_{out}$, the quadrotor exits the transition phase. It should be noted that $v_p$ is the predicted velocity using the WN\&CNets to steer the quadrotor from the hover state (waypoint 2) toward the following waypoints. 


The second scenario is 'opposite direction flight', where $(\mathbf{wp}_2-\mathbf{wp}_1)(\mathbf{wp}_3-\mathbf{wp}_2)<0$. In this scenario, it is the acceleration curve that presents a 'v' shape instead of the velocity. Thus, for the opposite-direction flight, similarly, we use the acceleration to determine the entrance and exit of the transition phase. Once $a < a_{in}$, the quadrotor enters the transition phase. And once $a > a_{out}$, the quadrotor exits the transition phase. Similarly,  $a_p$ is the predicted acceleration of the quadrotor using the WN\&CNets to steer the quadrotor from waypoint 2 toward waypoint 3. 
\begin{figure}[hbt]
    \centering
    \includegraphics[width=0.47\textwidth, trim = 200 200 230 150, clip]{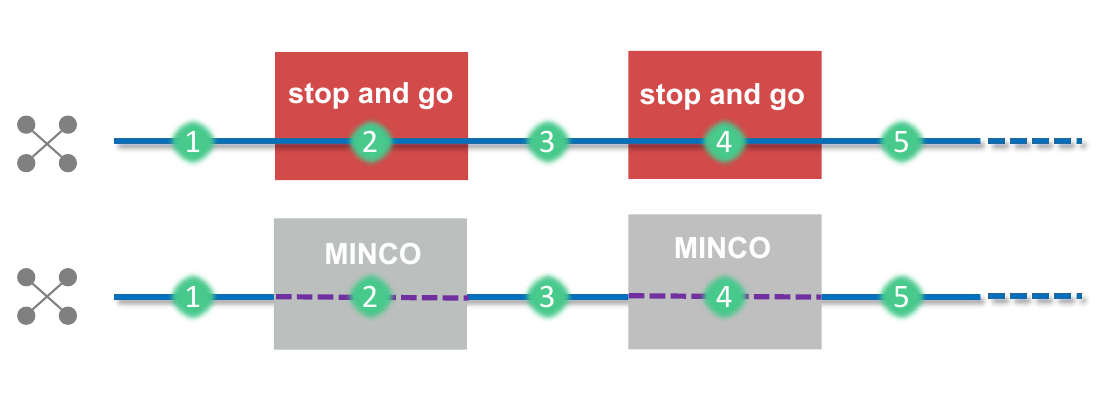}
    \caption{The sketch of the waypoint-switch strategy: with the proposed transition phase, the quadrotors don't have to hover at the waypoints. }
    \label{fig:london_tube_sketch}
\end{figure}

When the boundaries of the MINCO trajectories are determined (positions, velocities, and accelerations), the transition phase can be calculated. With the proposed transition phase, for multi-waypoint flight tasks, the quadrotor can switch the waypoints flexibly without unnecessary hover so that the quadrotor can fly with aggressive maneuvers with near-time-optimal performance (Fig. \ref{fig:london_tube_sketch} shows the sketch of the waypoint-switch strategy). 

\section{Experiment Result and Analysis}
In this section, we will analyze the time optimality and real-time capability of the proposed WN\&CNets through both simulation and real-world experiments. For simulation experiments, we will introduce CPC \cite{foehn2021time} as the benchmark and compare a great amount of CPC trajectories with the WN\&CNets for executing the same waypoint-constrained flight. For real-world experiments, we first use the WN\&CNets to navigate the quadrotor in two-waypoint flights with the first waypoint fixed or moving continuously. Next, we will demonstrate the performance of the WN\&CNets and MINCO transition through various multi-waypoint experiments.

\subsection{Experiment setup}

\begin{figure}[hbt]
    \centering
    \vspace{-0.2cm}
    \includegraphics[width=0.25\textwidth,trim = 0 0 0 0,clip]{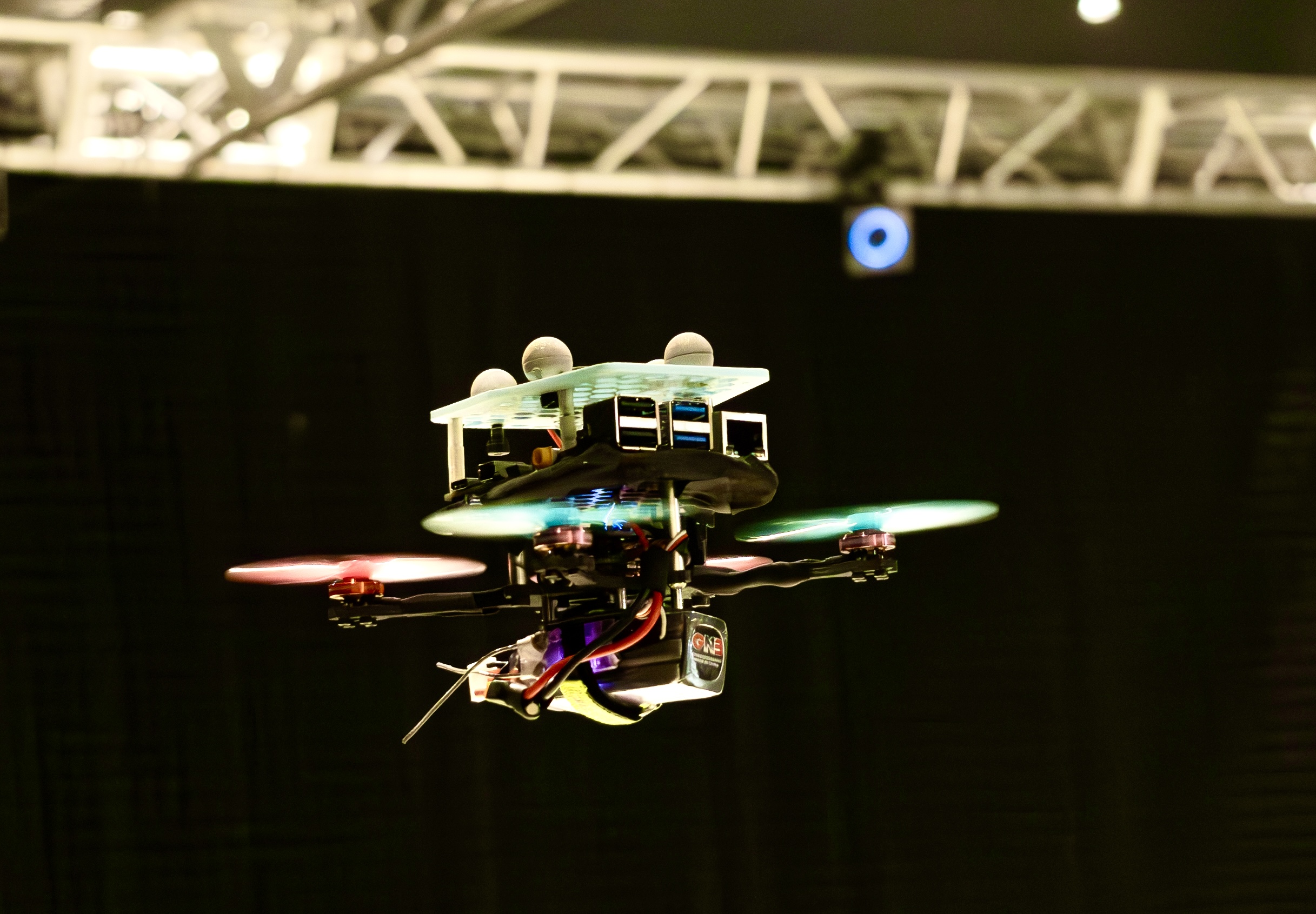}
    \caption{The quadrotor vehicle.}
    \label{fig: The quadrotor vehicle}
\end{figure}
\vspace{-0.25cm}

For real-world experiments, we utilize a self-made quadrotor as the flying platform, which weighs less than $200$g and boasts a thrust-to-weight ratio of about $2.2$, as shown in Fig. \ref{fig: The quadrotor vehicle}. To have precise and stable angular velocity tracking, the quadrotor is equipped with a $10$g autopilot with the Betaflight open-source software running onboard, which serves as an angular rate loop controller. The quadrotor is equipped with a Cool Pi computer with the proposed WN\&CNets running onboard. The rate commands and throttle commands are fed to the autopilot using the MAVLink protocol via the serial port. In our real-world experiments, to ensure safety and allow for a necessary margin, the free-flying space for the quadrotor to maneuver is set to be $5.5m \times 5.5m \times 2.0m$. 

To acquire accurate measurements of the quadrotor's states, the Optitrack motion capture system is used to provide accurate position and attitude measurements. To evaluate the delay of the measurements, we sum the delay caused by the motion capture system's calculation time which is provided by the Optitrack software and the delay of transmitting the data from the Optitrack system to the Cool Pi. The total delay is about $30ms$, which is considered in our network training.

On the Cool Pi, the WN\&CNet is deployed with LibTorch, and can achieve a maximum prediction frequency of $950$Hz. In our experiments, the quadrotor's position, velocity, attitude, and total thrust are sent to the WN\&CNets to predict the rate and throttle commands at $100$Hz. The predicted commands are sent to the autopilot to perform the maneuver. 

As for the transition phase (MINCO) and the benchmark (CPC), a classic differential flatness-based feed-forward and feedback strategy \cite{faessler2017differential} is employed.



\subsection{Time optimality of the WN\&CNets}
\begin{figure}[!h]
    \centering
\includegraphics[width=0.33\textwidth,trim = 350 70 330 70,clip]{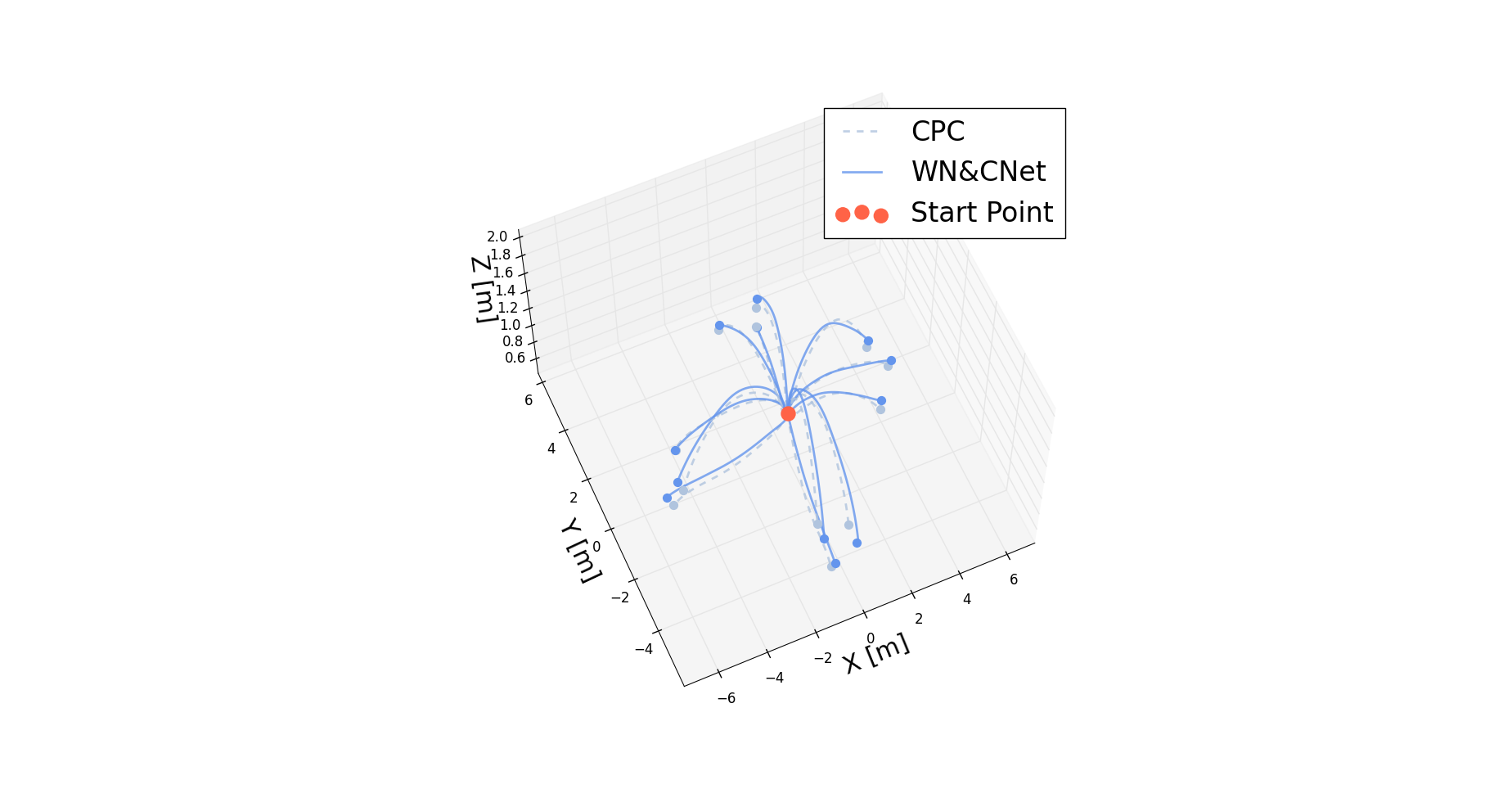}
    \caption{Part of 3D trajectories generated by the CPC and the WN\&CNets in simulation.}
    \label{fig:simulation_3d} 
\end{figure}
\begin{figure}[!h]
    \centering
    \vspace{-0.1cm}
    \includegraphics[width=0.49\textwidth,trim = 60 30 70 50,clip]{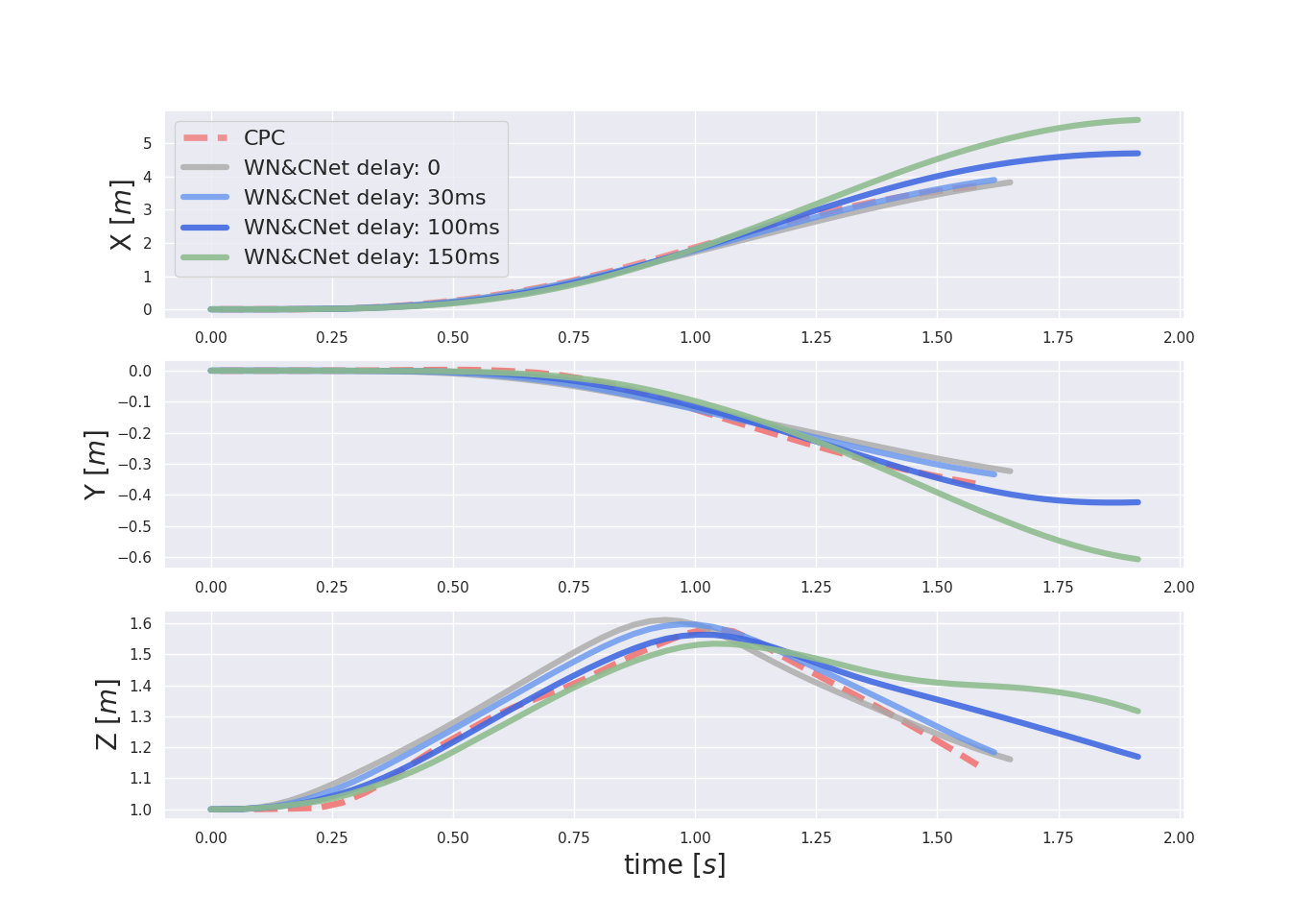}
     \caption{The navigation results of the WN\&CNets in simulations with different delay.}
    \label{fig:delay} 
\vspace{-0.4cm}
\end{figure}
To illustrate the time optimality of the WN\&CNets, we first compare the WN\&CNets with the CPC method in 1000 simple two-waypoint tasks distributed in the space of $[-5m,5m]\times[-5m,5m]\times[0.4m,2m]$. Part of the 3D trajectories generated by the CPC and the WN\&CNet are shown in Fig. \ref{fig:simulation_3d}. In the simulation, we utilize the dynamic model from Equation \ref{equ:quadrotor dynamics} as the quadrotors' model. The time step of the simulation is $0.025s$. The inputs for the simulation are the total thrust command $T_{c}$ and the torque command $\boldsymbol{\tau}_{c}$. The WN\&CNet provides the total thrust command $T_{c}$ and the angular rate command $\boldsymbol{\omega}_{c}$. We use a proportional control to calculate the torque commands $\boldsymbol{\tau}_{c}$ and send them to the dynamic model. As for the CPC, we directly use the trajectories generated by the CPC method.


We use a tolerance of $0.2m$ to determine if the quadrotor passes the waypoint. The simulation result shows that in these $1000$ flight tasks, the average time of the CPC trajectories is $1.89s$, while the average time for the quadrotor to complete the tasks navigated by the WN\&CNets is $1.91s$ (just one simulation step longer than CPC). The average minimum error in passing through the first waypoint by the CPC is $0.19m$ (just within the tolerance range), while by the WN\&CNets is $0.24m$. 

To illustrate the influence of the delay on flight performance, we introduce different delays in one of the aforementioned flights. The navigation results of the WN\&CNets with the ideal CPC are shown in Fig. \ref{fig:delay}. The quadrotor achieves the closest arrival time to the CPC in the simulation with a delay of $30ms$, which is the delay considered in the network training. It can also be found that with the larger delay mismatch, it is challenging for the same WN\&CNets to navigate the quadrotor.

\begin{figure}[!h]
    \centering
    \vspace{-0.2cm}
    \includegraphics[width=0.47\textwidth,trim = 100 20 70 30,clip]{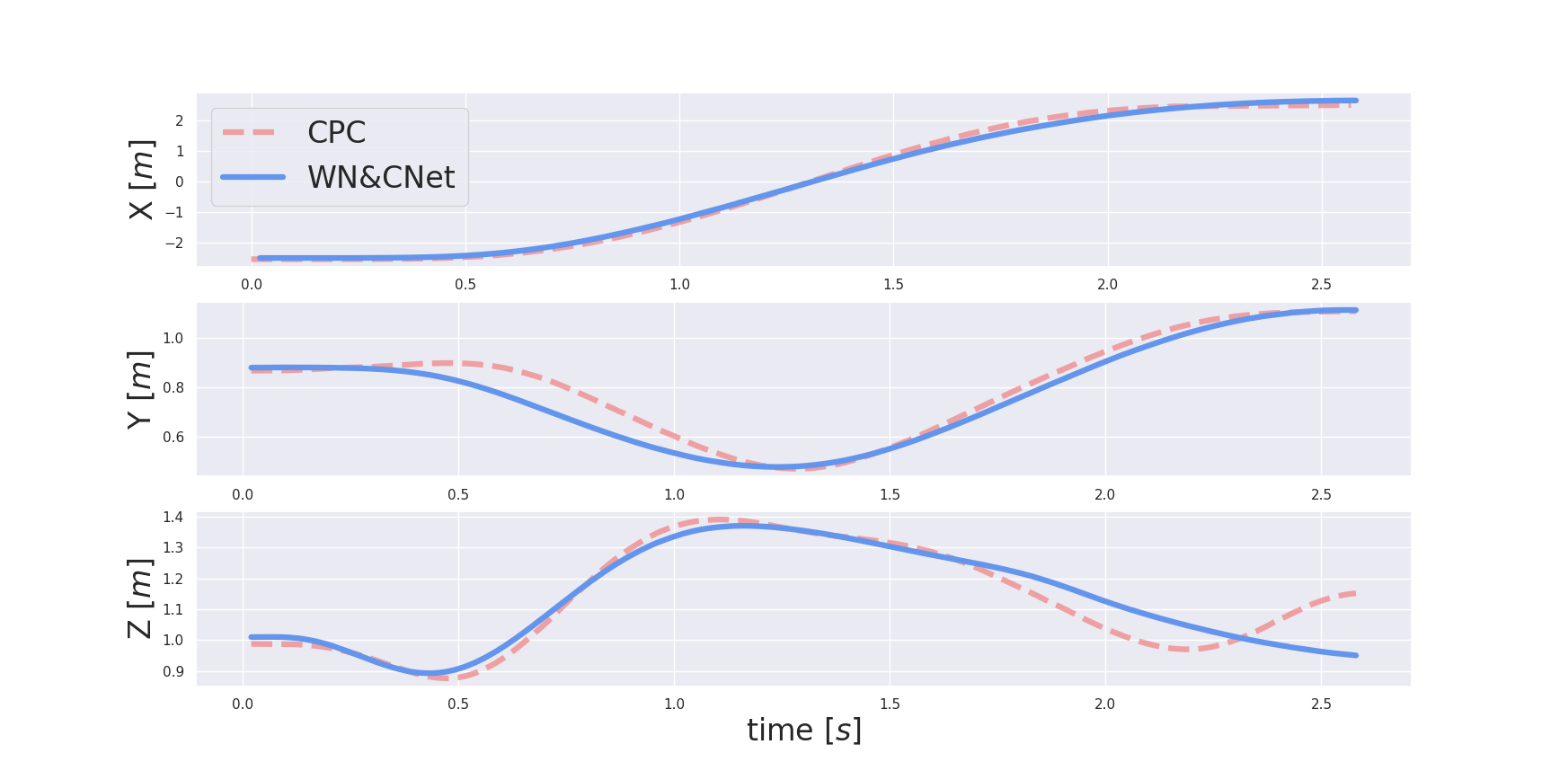}
    \vspace{-0.25cm}
    \caption{Comparison of the proposed method and the CPC method for a two-waypoint flight task in real-world experiments.}
    \label{fig:exp_single} 
    \vspace{-0.15cm}
\end{figure}
\begin{table*}[b]
\centering
\caption{The position of the waypoints and flight results }
\begin{tabular}{c c c c || c c c || c c c || c}
\hline
\multicolumn{1}{c}{Flight} &  \multicolumn{3}{c}{Start Point } & \multicolumn{3}{c}{Waypoint} & \multicolumn{3}{c}{Endpoint} & Maximum Error [m]\\ \hline
 & \multicolumn{1}{c}{$x$} & \multicolumn{1}{c}{$y$} & \multicolumn{1}{c||}{$z$} & \multicolumn{1}{c}{$x$} & \multicolumn{1}{c}{$y$} & \multicolumn{1}{c||}{$z$} & \multicolumn{1}{c}{$x$} & \multicolumn{1}{c}{$y$} & \multicolumn{1}{c||}{$z$} & \\ \hline
 $1$ &  $-2.5$ & $1.0$ & $1.0$ & $0.5\sin\frac{\pi}{5}t$ & $0.5\cos\frac{\pi}{5}t + 1.5$ & $1.5$ & $2.5$ & $1.0$ & $1.0$ & $0.39$ \\ \hline

 $2$ & $2.5$ & $1.0$ & $1.0$ & $0.5\sin(\frac{\pi}{5}t+1.0)$ & $0.5\cos(\frac{\pi}{5}t+1.0)$ & $1.0$ & $-2.5$ & $-1.0$ & $1.0$ & $0.29$ \\ \hline

 $3$ & $-2.5$ & $-1.0$ & $1.0$ &  $0.5\sin(\frac{\pi}{5}t+1.5)$ & $0.5\cos(\frac{\pi}{5}t+1.5)-1.5$ & $0.5$ & $2.5$ & $-1.0$ & $1.0$ & $0.37$ \\ \hline 

$4$ & $2.5$ & $-1.0$ & $1.0$ & $0.5\sin(\frac{\pi}{5}t+1.0)$ & $0.5\cos(\frac{\pi}{5}t+1.0)$ & $1.0$ & $-2.5$ & $1.0$ & $1.0$ & $0.25$ \\ \hline
\end{tabular}
\label{tab: waypoints detail}
\end{table*}
\begin{figure*}[b]
    \vspace{-0.35cm}
    \hspace{10mm}
    \subfigure[The scenarios of the quadrotor passing through the moving waypoints for 4 flights. We use blue, green and red gates to represent the moving waypoints.]{\includegraphics[width=0.38\textwidth,trim = 80 140 750 200, clip]{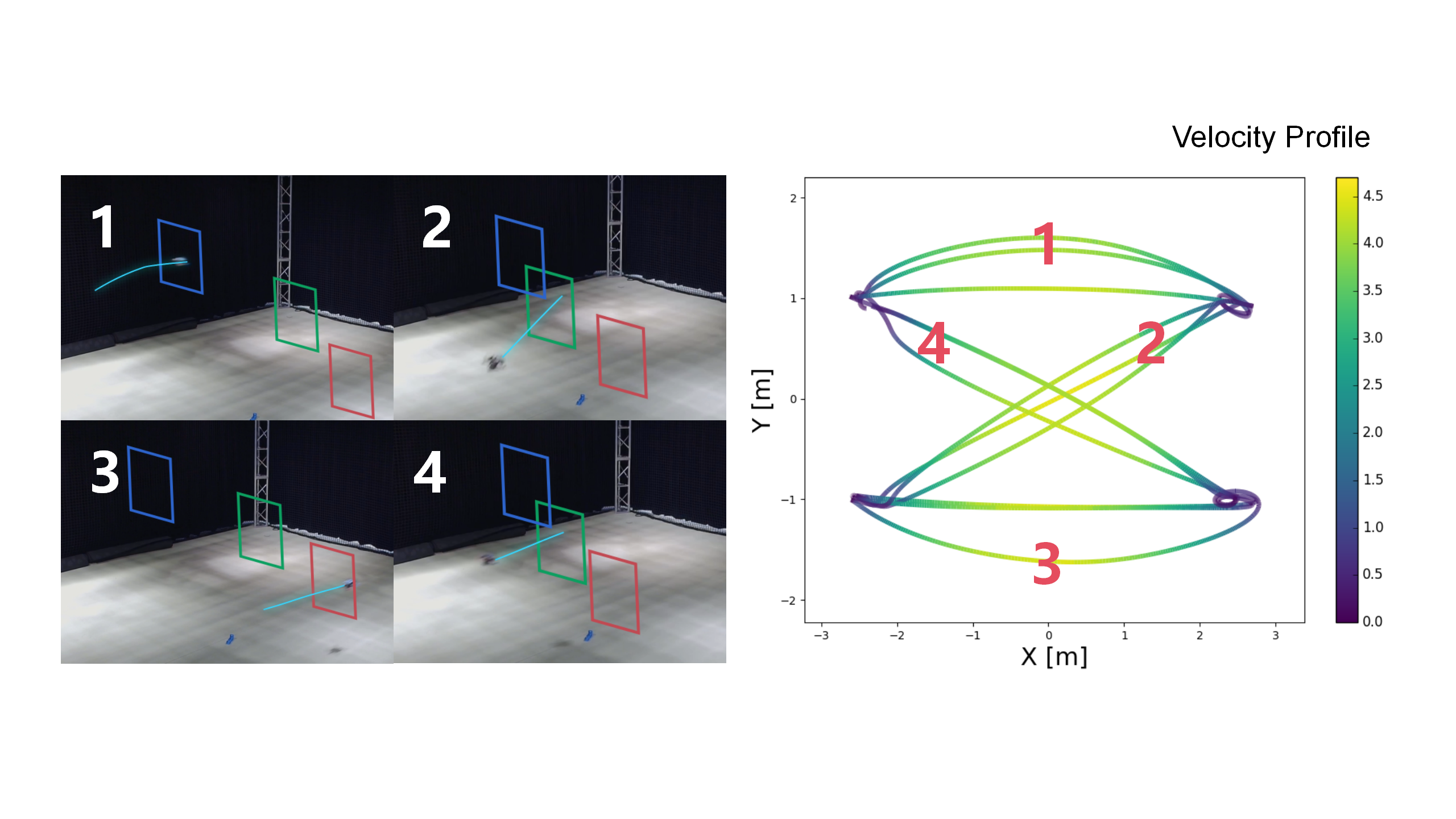}}
    \hspace{15mm}
    \subfigure[The trajectories and velocity profiles of the quadrotor maneuvering for $3$ laps.]{\includegraphics[width=0.38\textwidth,trim = 780 150 50 100, clip]{Figures/exp1_3.png}} 
    \hspace{-25mm}
    \caption{The scenarios and the trajectories of the quadrotor flying through moving waypoints navigated using the WN\&CNets.} 
    \label{fig:moving}
\end{figure*}

We also perform the comparison of the WN\&CNets and CPC in real-world flights. Fig. \ref{fig:exp_single} demonstrates the position curves generated by the CPC method and the WN\&CNets in one specific two-waypoint task, in which the start point is $(-2.55, 1.0, 1.0)$ and the waypoints are first $(0.0, 0.25, 1.4)$ and then $(2.5, 1.1, 1.0)$. The result indicates that for the same flight task, the trajectories of the WN\&CNet are very close to the one generated by the CPC method, while achieving real-time computing. It indicates that the proposed method achieves the online near-time-optimal performance.

The second real-world experiment is designed to let the quadrotor fly through moving waypoints to demonstrate that the proposed method can handle dynamic waypoints online. Table \ref{tab: waypoints detail} lists the positions of the start points, moving waypoints and endpoints. In this experiment, the quadrotor will first hover at the start points and then use the WN\&CNets to fly through the continuously moving waypoints and finally hover at the endpoints. During the flight, the quadrotor only knows the current position of the waypoints instead of their whole trajectories. The positions of the waypoints are fed to the WN\&CNets at the frequency of $100$Hz. In this experiment, the quadrotor maneuvers in this manner for $3$ laps. The maximum errors of all the laps between the quadrotor and the moving waypoints are also listed in Table \ref{tab: waypoints detail}. The experiment scenarios and trajectories are presented in Fig. \ref{fig:moving}. This experiment indicates one important advantage over the benchmark, CPC, that is due to the neural networks' high efficiency and high-frequency prediction, the proposed method is able to handle the dynamic waypoints online while the CPC method needs several seconds to several minutes to optimize a feasible trajectory which makes it impossible to run online and handle dynamic waypoints during the flight.  

\subsection{WN\&CNets and transition strategy for multi-waypoint flight tasks}

\begin{figure}[hbt]
    \centering
    \vspace{-2mm}
   \includegraphics[width=0.3\textwidth,trim = 100 100 820 150,clip]{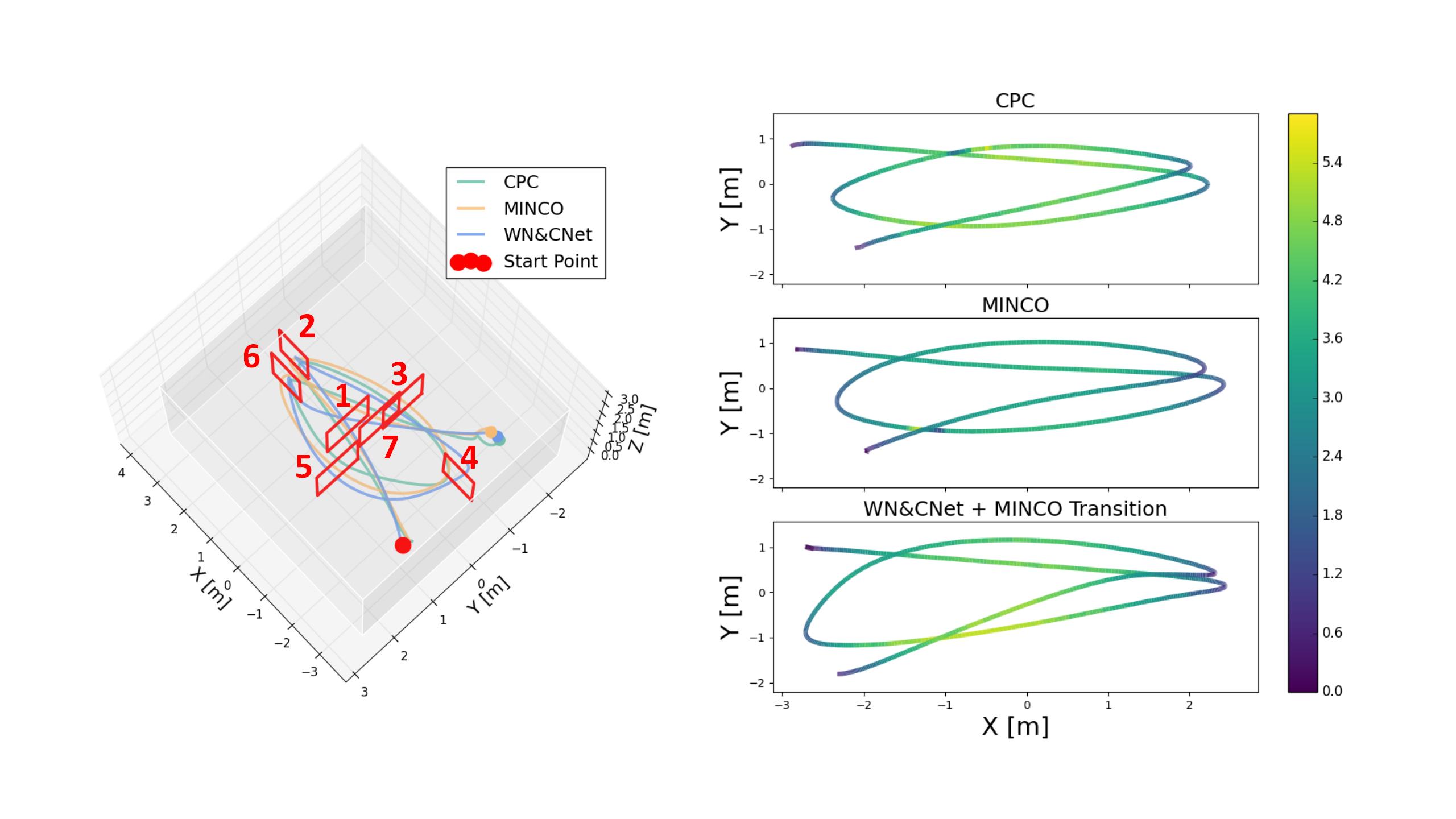}
    \vspace{-3mm}
    \caption{Trajectories of the quadrotor flying through $7$ waypoints via WN\&CNets and transition phase strategy along with the benchmarks. The red numbers represent the order of the waypoints.}
    \label{fig: real-m}
\end{figure}

In this section, we design an experiment to test the proposed transition strategy. In this experiment, the quadrotor first hovers at the start point, then flies through $7$ waypoints aggressively and finally hovers at the endpoint (Fig. \ref{fig: real-m}). The positions of each waypoint are listed in Table \ref{tab: real-world-m waypoints}.
\begin{table}[h]
\vspace{-2mm}
\setlength{\belowdisplayskip}{0pt}
\caption{Waypoints' position for multi-waypoint experiment}
\label{table_example}
\begin{center}
\begin{tabular}{c|c|c}
\Xhline{1px}
Start Point & $wp_1$  & $wp_2$ \\ \hline
$(-2.8,1.0,1.0)$ & $(0.0, 0.4, 1.9)$ & $(2.5, 0.0, 1.0)$ \\ \Xhline{1px}
$wp_3$ & $wp_4$ & $wp_5$ \\ \hline
$(0.0, -1.0, 0.4)$ & $(-2.5, -0.6, 1.0)$ & $(-0.5, 1.0, 1.5)$ \\ \Xhline{1px}
$wp_6$  & $wp_7$ & Endpoint\\ \hline
$(2.2, 0.4, 1.0)$ & $(0.0, -0.4, 0.6)$ & $(-2.2, -1.5, 1.0)$ \\ \Xhline{1px}
\end{tabular}
\end{center}
\label{tab: real-world-m waypoints}
\end{table}
\setlength{\tabcolsep}{3pt}

\begin{figure}[!h]
    \centering
    \includegraphics[width=0.5\textwidth,trim = 100 0 70 80,clip]{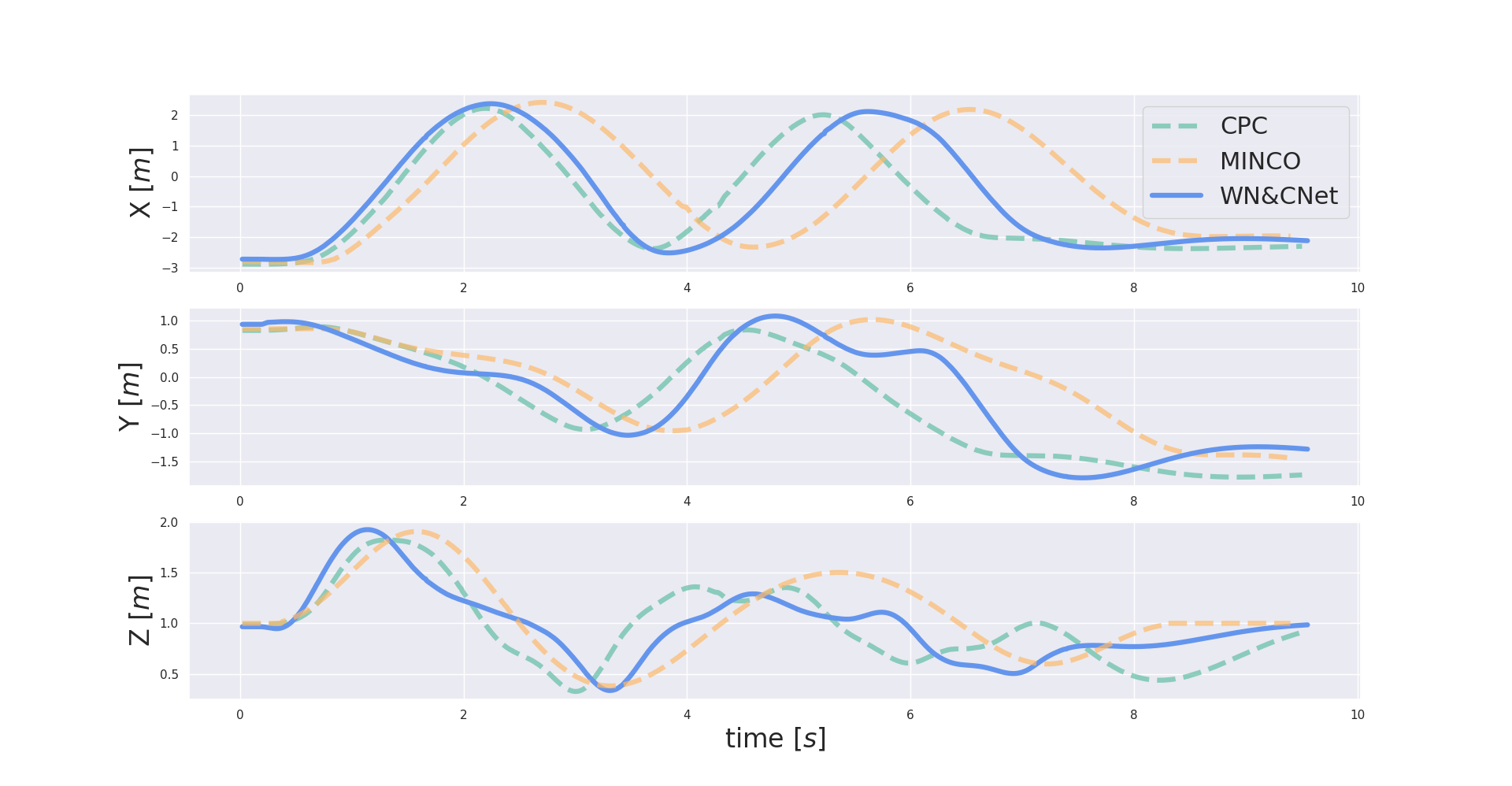}
    \vspace{-6mm}
     \caption{Comparison of the proposed method, CPC and MINCO for a multi-waypoint flight task in real-world experiments.}
    \label{fig: exp2-time} 
\end{figure}

At the same time, we use the CPC method and the MINCO method to steer the quadrotor to fly the same track as the benchmarks. In the experiments, the CPC, the MINCO and our WN\&CNets can navigate the quadrotor to fly through all the waypoints successfully and the quadrotor can achieve a maximum speed of up to $5.0m/s$. The trajectory curves generated by the three methods are shown in Fig. \ref{fig: exp2-time}. To compare the performance of the three methods, we list the arrival time $t_f$, the maximum velocity $v_{max}$, the maximum error $E_{max}$ of passing through the waypoints, and the processing time $T_p$ of generating the optimal control commands in Table \ref{tab: m results}.

From Table 3, it can be seen that although the arrival time of the proposed method ($7.09s$) is slightly longer than the CPC method ($6.66s$), the processing time of the proposed method and the CPC method differs by orders of magnitude. The CPC method needs $9min$ to calculate the time-optimal trajectories while the proposed method can be run at the frequency of $100$Hz (the WN\&CNets can achieve a prediction frequency of $950$Hz, but $100$Hz is the frequency that is used in our experiments) to predict the time-optimal trajectories. In terms of the tracking error, due to the tolerance of $0.2m$ in the optimization problem as well as the presence of unknown disturbances and model inaccuracy, the CPC trajectory, which can not replan online, passes through waypoints with maximum error as $0.29m$.
\begin{table}[h]
\setlength{\belowdisplayskip}{0pt}
\vspace{2mm}
\caption{Experiment results: comparison between each method }
\label{table_example}
\begin{center}
\begin{tabular}{c|c|c|c|c}
\hline
  & $t_f$ [s] & $v_{max}$ [m/s] & $E_{max}$ [m] & $T_p$ \\ \hline
CPC & $6.66$ & $5.8$ & $0.29$ & $9min$ \\ \hline
MINCO & $8.20$ & $5.0$ & $0.18$ & $\leq 0.01s$ \\ \hline
WN\&CNet & $7.09$ & $5.6$ & $0.22$ & $\leq 0.01s$ \\ \hline
\end{tabular}
\end{center}
\label{tab: m results}
\vspace{-3mm}
\end{table}
\setlength{\tabcolsep}{3pt}

The MINCO can be solved online and analytically passes through each waypoint accurately, which means that the errors are completely caused by trajectory tracking errors. The maximum error of the WN\&CNets along with the MINCO transition is $0.22m$, which is slightly larger than the tolerance for generating the time optimal datasets while it is close to the average error in the simulation. To conclude, in multi-waypoint flight tasks, although the proposed method slightly loses time optimality, it significantly surpasses the CPC method in its efficiency which makes it more practical and flexible.

\begin{figure}[hbt]
    \centering
    \vspace{-2mm}
    \includegraphics[width=0.38\textwidth,trim=300 80 380 60,clip]{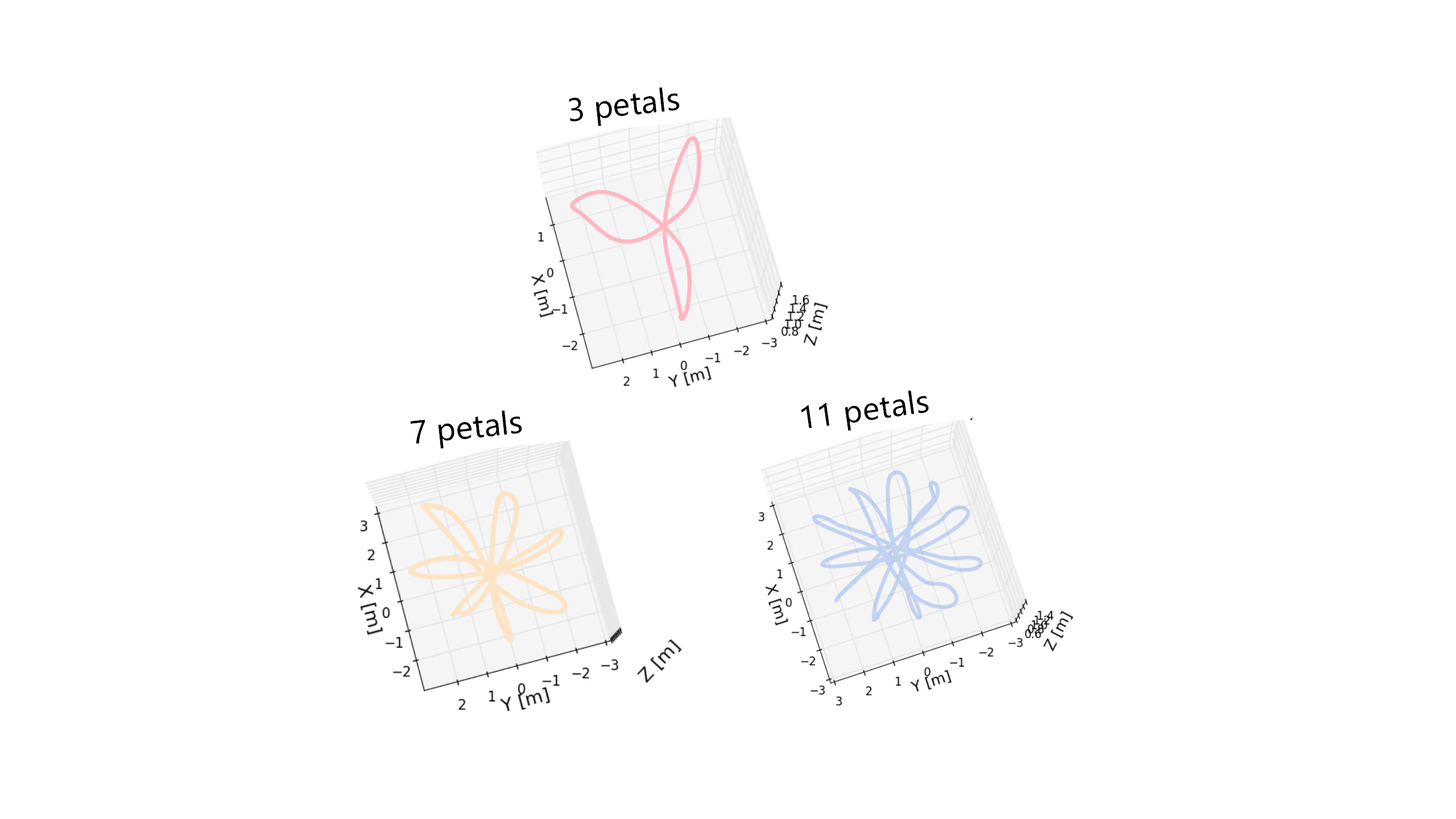}
    \vspace{-2mm}
    \caption{Flower-like trajectories generated by the proposed method (a lap of 11-petal trajectory contains 22 waypoints).}
    \label{fig: flower}
    \vspace{-2mm}
    \end{figure}

To illustrate the feasibility of the proposed method to be applied to a wider range of scenarios, we also demonstrate that the quadrotor can fly with flower-like trajectories (Fig. \ref{fig: flower}) and continuously pass through moving target points without hovering (Fig. \ref{fig: exp4}).
    
\subsection{WN\&CNets with headings}
\begin{figure}[hbt]
    \centering
    \includegraphics[width=0.48\textwidth,trim=190 230 190 280, clip]{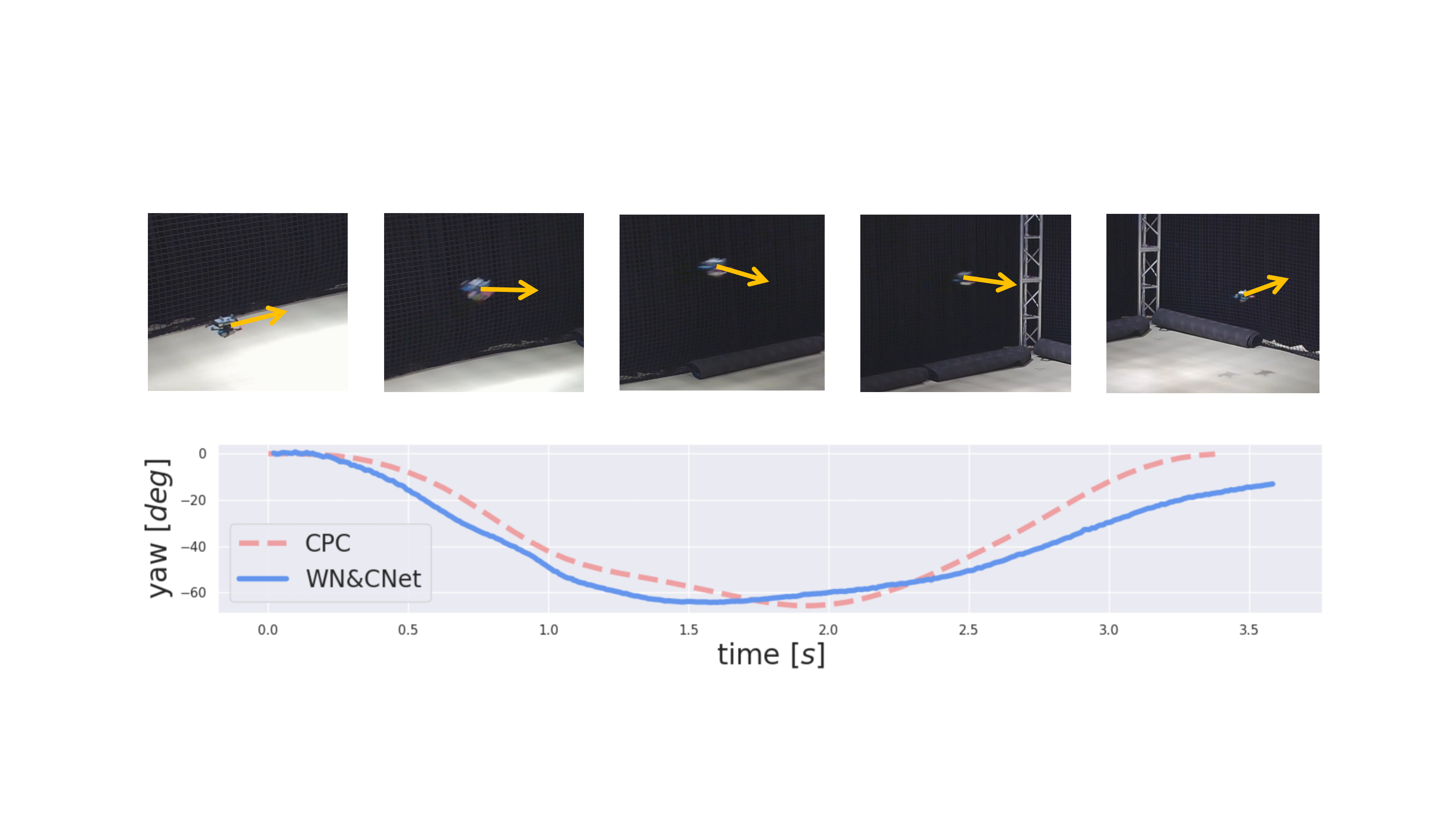}
    \vspace{-3mm}
    \caption{The quadrotor is steered by the WN\&CNets (compared with the CPC) to pass through the first waypoint with a heading of $-60 ^\circ$ and hover at the final waypoint with a heading of $-0 ^\circ$ (the yellow arrow indicates the heading of the quadrotor).}
    \label{fig: exp5}
    \vspace{-3mm}
\end{figure}

As we discussed in Section \ref{sec:optimization problem}, for some flight tasks, the heading needs to be specified at the waypoints. Although in this letter, we mainly focus on time-optimal maneuvers with free heading, we also generate the dataset using constraint (\ref{cpc:yaw}) and train another WN\&CNets with heading constraints. In terms of training, the input heading measurement of the network should be changed into a relative heading to the desired one, which is similar to the relative position inputs. The flying results show that the WN\&CNets with headings can steer the quadrotor pass through the waypoints with specified headings in the minimum time. The pictures and the yaw curves of the experiments are shown in Fig. \ref{fig: exp5}.

\section{CONCLUSIONS}
In this letter, we develop a novel network WN\&CNet which can navigate the quadrotor to fly through multiple waypoints with the minimum time. Due to the quadrotors' high-dimension state space and very limited time-optimal training data, the trained neural network can only guide the quadrotor fly through $2$ waypoints and hover at the second waypoint. To address this issue, with the aid of MINCO trajectories, we develop a transition phase strategy to help the quadrotor jump over the so-called stop-and-go phase. So that the quadrotor can consecutively fly through multiple waypoints with its aggressive maneuverability. We also conduct simulations and real-world experiments to test our method. The results indicate that although the proposed method slightly loses the time-optimality compared to the CPC method, it gains the ability to be run online with $100$Hz in return, which makes it more practical for real-world flying tasks that need aggressive maneuvers.

There are several directions for future work. The real inputs of the quadrotors, the rotors' speed, can be added to the quadrotor's model and the optimization problem to further improve the time optimality. The sensors’ delay would be considered in our future work so that they can be estimated and compensated online to eliminate the negative effects on the network's prediction accuracy. More computationally efficient time-optimal trajectory generation methods can be adopted to replace the CPC to generate datasets, so that we can acquire much more training data in confined time and expand the scope of application of the WN\&CNets. Two-drone racing with the proposed WN\&CNets can also be an interesting topic to research on. 


\begin{thebibliography}{10}

\bibitem{hanover2024autonomous}
D.~Hanover, A.~Loquercio, L.~Bauersfeld, A.~Romero, R.~Penicka, Y.~Song, G.~Cioffi, E.~Kaufmann, and D.~Scaramuzza, ``Autonomous drone racing: A survey,'' {\em IEEE Transactions on Robotics}, 2024.

\bibitem{zhou2019robust}
B.~Zhou, F.~Gao, L.~Wang, C.~Liu, and S.~Shen, ``Robust and efficient quadrotor trajectory generation for fast autonomous flight,'' {\em IEEE Robotics and Automation Letters}, vol.~4, no.~4, pp.~3529--3536, 2019.

\bibitem{ren2022bubble}
Y.~Ren, F.~Zhu, W.~Liu, Z.~Wang, Y.~Lin, F.~Gao, and F.~Zhang, ``Bubble planner: Planning high-speed smooth quadrotor trajectories using receding corridors,'' in {\em 2022 IEEE/RSJ International Conference on Intelligent Robots and Systems (IROS)}, pp.~6332--6339, IEEE, 2022.

\bibitem{foehn2022alphapilot}
P.~Foehn, D.~Brescianini, E.~Kaufmann, T.~Cieslewski, M.~Gehrig, M.~Muglikar, and D.~Scaramuzza, ``Alphapilot: Autonomous drone racing,'' {\em Autonomous Robots}, vol.~46, no.~1, pp.~307--320, 2022.

\bibitem{cui2016search}
J.~Q. Cui, S.~K. Phang, K.~Z. Ang, F.~Wang, X.~Dong, Y.~Ke, S.~Lai, K.~Li, X.~Li, J.~Lin, {\em et~al.}, ``Search and rescue using multiple drones in post-disaster situation,'' {\em Unmanned Systems}, vol.~4, no.~01, pp.~83--96, 2016.

\bibitem{mellinger2011minimum}
D.~Mellinger and V.~Kumar, ``Minimum snap trajectory generation and control for quadrotors,'' in {\em 2011 IEEE International Conference on Robotics and Automation}, pp.~2520--2525, IEEE, 2011.

\bibitem{bry2015aggressive}
A.~Bry, C.~Richter, A.~Bachrach, and N.~Roy, ``Aggressive flight of fixed-wing and quadrotor aircraft in dense indoor environments,'' {\em The International Journal of Robotics Research}, vol.~34, no.~7, pp.~969--1002, 2015.

\bibitem{faessler2017differential}
M.~Faessler, A.~Franchi, and D.~Scaramuzza, ``Differential flatness of quadrotor dynamics subject to rotor drag for accurate tracking of high-speed trajectories,'' {\em IEEE Robotics and Automation Letters}, vol.~3, no.~2, pp.~620--626, 2017.

\bibitem{webb2013kinodynamic}
D.~J. Webb and J.~Van Den~Berg, ``Kinodynamic rrt*: Asymptotically optimal motion planning for robots with linear dynamics,'' in {\em 2013 IEEE international conference on robotics and automation}, pp.~5054--5061, IEEE, 2013.

\bibitem{liu2018search}
S.~Liu, K.~Mohta, N.~Atanasov, and V.~Kumar, ``Search-based motion planning for aggressive flight in se (3),'' {\em IEEE Robotics and Automation Letters}, vol.~3, no.~3, pp.~2439--2446, 2018.

\bibitem{foehn2021time}
P.~Foehn, A.~Romero, and D.~Scaramuzza, ``Time-optimal planning for quadrotor waypoint flight,'' {\em Science Robotics}, vol.~6, no.~56, p.~eabh1221, 2021.

\bibitem{shen2023aggressive}
Y.~Shen, J.~Zhou, D.~Xu, F.~Zhao, J.~Xu, J.~Chen, and S.~Li, ``Aggressive trajectory generation for a swarm of autonomous racing drones,'' in {\em 2023 IEEE/RSJ International Conference on Intelligent Robots and Systems (IROS)}, pp.~7436--7441, IEEE, 2023.

\bibitem{wang2022geometrically}
Z.~Wang, X.~Zhou, C.~Xu, and F.~Gao, ``Geometrically constrained trajectory optimization for multicopters,'' {\em IEEE Transactions on Robotics}, vol.~38, no.~5, pp.~3259--3278, 2022.

\bibitem{fork2023euclidean}
T.~Fork and F.~Borrelli, ``Euclidean and non-euclidean trajectory optimization approaches for quadrotor racing,'' {\em arXiv preprint arXiv:2309.07262}, 2023.

\bibitem{qin2023time}
C.~Qin, M.~S. Michet, J.~Chen, and H.~H.-T. Liu, ``Time-optimal gate-traversing planner for autonomous drone racing,'' {\em arXiv preprint arXiv:2309.06837}, 2023.

\bibitem{romero2022model}
A.~Romero, S.~Sun, P.~Foehn, and D.~Scaramuzza, ``Model predictive contouring control for time-optimal quadrotor flight,'' {\em IEEE Transactions on Robotics}, vol.~38, no.~6, pp.~3340--3356, 2022.

\bibitem{romero2022time}
A.~Romero, R.~Penicka, and D.~Scaramuzza, ``Time-optimal online replanning for agile quadrotor flight,'' {\em IEEE Robotics and Automation Letters}, vol.~7, no.~3, pp.~7730--7737, 2022.

\bibitem{song2023reaching}
Y.~Song, A.~Romero, M.~M{\"u}ller, V.~Koltun, and D.~Scaramuzza, ``Reaching the limit in autonomous racing: Optimal control versus reinforcement learning,'' {\em Science Robotics}, vol.~8, no.~82, p.~eadg1462, 2023.

\bibitem{kaufmann2023champion}
E.~Kaufmann, L.~Bauersfeld, A.~Loquercio, M.~M{\"u}ller, V.~Koltun, and D.~Scaramuzza, ``Champion-level drone racing using deep reinforcement learning,'' {\em Nature}, vol.~620, no.~7976, pp.~982--987, 2023.

\bibitem{tang2018learning}
G.~Tang, W.~Sun, and K.~Hauser, ``Learning trajectories for real- time optimal control of quadrotors,'' in {\em 2018 IEEE/RSJ International Conference on Intelligent Robots and Systems (IROS)}, pp.~3620--3625, 2018.

\bibitem{izzo2020real}
D.~Izzo and E.~{\"O}zt{\"u}rk, ``Real-time optimal guidance and control for interplanetary transfers using deep networks,'' {\em arXiv preprint arXiv:2002.09063}, 2020.

\bibitem{li2020aggressive}
S.~Li, E.~{\"O}zt{\"u}rk, C.~De~Wagter, G.~C. De~Croon, and D.~Izzo, ``Aggressive online control of a quadrotor via deep network representations of optimality principles,'' in {\em 2020 IEEE International Conference on Robotics and Automation (ICRA)}, pp.~6282--6287, IEEE, 2020.

\bibitem{origer2023guidance}
S.~Origer, C.~De~Wagter, R.~Ferede, G.~C. de~Croon, and D.~Izzo, ``Guidance \& control networks for time-optimal quadcopter flight,'' {\em arXiv preprint arXiv:2305.02705}, 2023.

\end{thebibliography}
\bibliographystyle{ieeetr}

\end{document}